\documentclass[letterpaper, 10 pt, conference]{ieeeconf}  

\IEEEoverridecommandlockouts                              

\usepackage{cite}
\usepackage{amsmath,amssymb,amsfonts}
\usepackage{algorithmic}
\usepackage{graphicx}
\usepackage{textcomp}
\usepackage{xcolor}
\usepackage{subfigure}
\usepackage{tablefootnote}
\UseRawInputEncoding

\title{Lane-GNN: Integrating GNN for Predicting Drivers' Lane Change Intention}

\author{Hongde Wu and Mingming Liu
	\thanks{H. Wu and M. Liu are with the School of Electronic Engineering, Dublin City University, Dublin, Ireland. M. Liu is also with the Insight SFI Research Centre for Data Analytics, Dublin City University, Ireland. Corresponding author's email: {\tt mingming.liu@dcu.ie}}
}

\begin{document}

	\maketitle
	\thispagestyle{empty}
	\pagestyle{empty}

	\begin{abstract}
		
		Nowadays, intelligent highway traffic network is playing an important role in modern transportation infrastructures. A variable speed limit (VSL) system can be facilitated in the highway traffic network to provide useful and dynamic speed limit information for drivers to travel with enhanced safety. Such system is usually designed with a steady advisory speed in mind so that traffic can move smoothly when drivers follow the speed, rather than speeding up whenever there is a gap and slowing down at congestion. However, little attention has been given to the research of vehicles' behaviours when drivers left the road network governed by a VSL system, which may largely involve unexpected acceleration, deceleration and frequent lane changes, resulting in chaos for the subsequent highway road users. In this paper, we focus on the detection of traffic flow anomaly due to drivers' lane change intention on the highway traffic networks after a VSL system. More specifically, we apply graph modelling on the traffic flow data generated by a popular mobility simulator, SUMO, at road segment levels. We then evaluate the performance of lane changing detection using the proposed Lane-GNN scheme, an attention temporal graph convolutional neural network, and compare its performance with a temporal convolutional neural network (TCNN) as our baseline. Our experimental results show that the proposed Lane-GNN can detect drivers' lane change intention within 90 seconds with an accuracy of $99.42\%$ under certain assumptions. Finally, some interpretation methods are applied to the trained models with a view to further illustrate our findings.
		
		%
		%
		
	\end{abstract}

	\section{Introduction}
	
\begin{figure}[htp]
		\centering
		\includegraphics[width=7.5 cm]{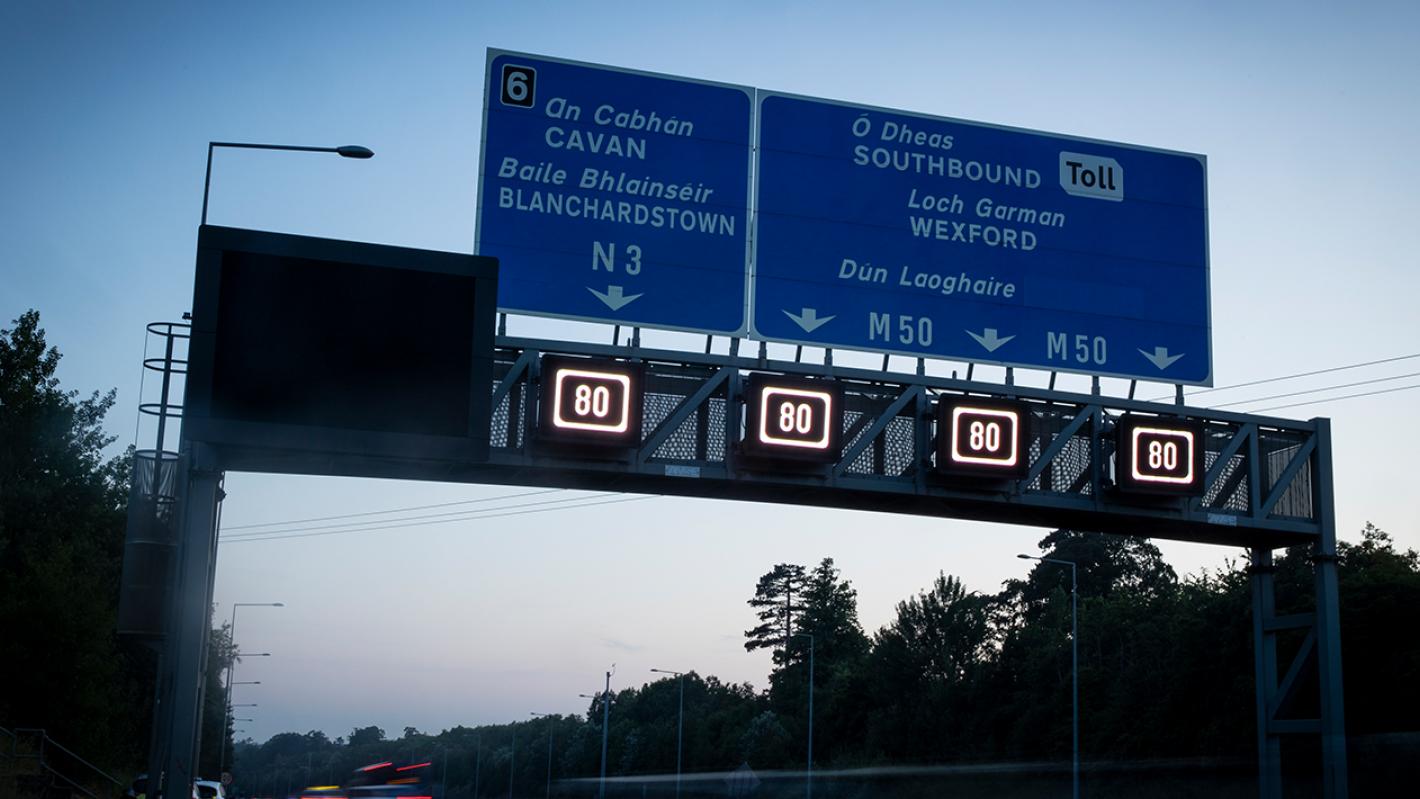}
		\caption{Variable speed limits (80km/h) are shown for four lanes of the M50 highway traffic networks in Dublin city, Ireland \protect\footnote[1]{https://www.rod.ie/projects/enhancing-motorway-operation-services}.}
		\label{fig_sas}
	\end{figure}

	At present, Intelligent Transportation Systems (ITS) is playing an important role in addressing various challenges to maximize its impact on safety and sustainability for citizens travelling in cities \cite{aldegheishem2018smart}. A road-side camera monitoring system, for instance, can be deployed in a highway network to track moving vehicles by leveraging advanced computer vision techniques and modern telecommunication networks. Useful information including averaged speed and density of vehicles on a given stretch of road can be obtained using big data analysis and artificial intelligence techniques. Real-time data and insight can be exchanged to city stakeholders through different channels using dedicated application programming interfaces (APIs) \cite{mejia2021vehicle}. In this context, a variable speed limit (VSL) system can be deployed and integrated with intelligent infrastructures to alleviate traffic congestion and maximize traffic flows in different scenarios \cite{kuvsic2020extended, liu2021mpc}. For instance, Fig. \ref{fig_sas} illustrates a VSL system currently deployed in the highway network (M50) in Dublin, Ireland. The VSL system is designed in a way that vehicles on each lane of the road network can drive at a safe speed in km/h shown below the overhead gantry signboard \footnote[1]{https://www.rod.ie/projects/enhancing-motorway-operation-services}. It is worth noting that although driving at a lower speed below the VSL is legal, it is usually not recommended as it will easily lead to a bottleneck for subsequent vehicles trying to follow the speed, and thus reducing traffic flow \footnote[2]{https://www.theaa.ie/blog/m50-variable-speed-limits/}.
	
	A large body of work has been found in the literature for an optimal design of advisory speed for different application purposes, see papers \cite{gu2018design, chen2021intelligent, liu2015topics,7350149, gu2014optimised, liu2015intelligent, tal2015vehicular, xiang2015closed} in this direction. However, little attention has been given for the study of drivers' behaviours after the road segments with speed advisory, where a driver may intend to drive freely with acceleration/de-acceleration, frequent lane-changing according to a psychological study in \cite{jeon2014effects}. More specifically, Fig. \ref{fig_highway} illustrates the M50 highway traffic network, where the VSL system is implemented on the segment in green and a subsequent segment in red without a VSL system. This implies that once vehicles leave the road segment in green, it is more likely that vehicles may drive freely and change lane frequently on the road segment in red. Thus, it is important to understand how likely drivers will change lanes by using the observable information collected from the road networks. This is the exact research challenge to be addressed in this work. With this in mind, our main contributions of this paper can be summarised as follows:
	
	\begin{figure}[htp]
		\centering
		\includegraphics[width=7.0 cm]{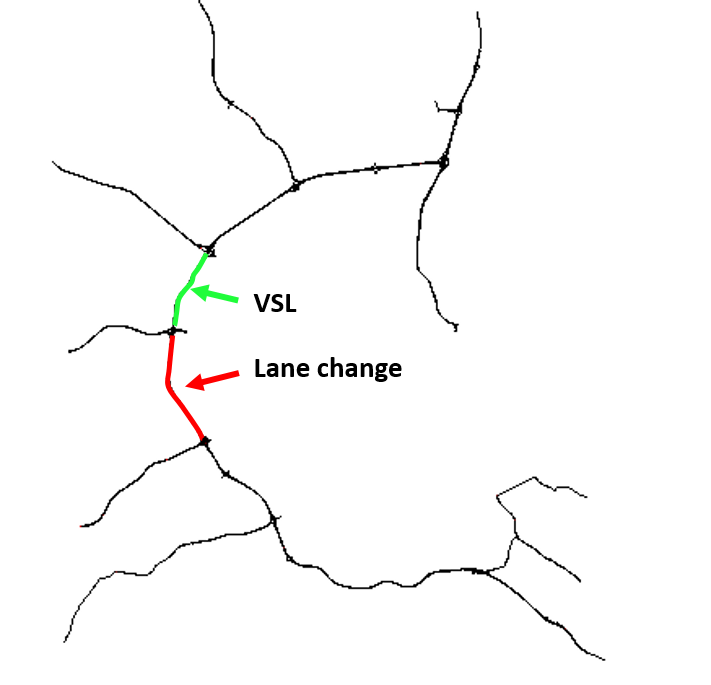}
		\caption{The simulated framework of the M50 highway network in Dublin city using SUMO. In the real world, the segment in green is implemented with VSL and the segment in red indicates the area in which frequent lane changing behaviours may happen. }
		\label{fig_highway}
	\end{figure}

    	\begin{itemize}
    	   	
    	\item [1.] We model highway traffic networks using graphs with lanes as nodes and connectivity between lanes as edges to extract the graph features for lane changing detection. Given this setup, we propose Lane-GNN, an attention temporal graph convolutional neural network, to predict drivers' lane changing intention.  
    	
    	\item [2.] We evaluate the performance of our proposed scheme using data generated from a mobility simulator namely SUMO \cite{behrisch2011sumo}. The performance of the scheme is compared with a solution based on temporal convolutional neural network (TCNN) as our baseline. In comparison with the baseline algorithm, our proposed Lane-GNN method shows superiority under certain assumptions. 
    
    	\item [3.] Temporal and spectral domain analysis, including standard deviation and spectral information divergence for features, are carried out to further interpret our models and findings. 
    	
    \end{itemize}
	
	The remaining parts of the paper are organized as follows. We review some deep learning based analysis for traffic flow and lane change detection in Section \ref{relatework}. The experiment design, data processing and neural network architecture are demonstrated in Section \ref{methodology}. Experimental results and further details regarding to the results are discussed in Section \ref{experiment}. Finally, we conclude our work in Section \ref{conclusion}.

	\section{Related works} \label{relatework}

      Many works have been found in the literature using deep learning methods for traffic flow analysis. More recently, deep belief networks \cite{huang2014deep}, autoencoder\cite{lv2014traffic} and recurrent neural network (RNN) based approaches \cite{tian2018lstm} have been implemented to analyze the sequential traffic flow data leveraging the long term temporal dependencies. Jointly working with sequential deep learning models, by segmenting the city into multiple areas and grids, CNN architectures with temporal units have been devised to access both spatial and temporal information where the traffic flow is processed into sequential 2-D data \cite{8526506} \cite{ma2017learning}. However, the above methods meet with common limitations for traffic flow analysis since they neglect the natural non-Euclidean property (e.g., graph) in road networks. Also, some previous works have shown the effectiveness of lane change detection, using HMM \cite{li2016lane} and LSTM based methods \cite{tang2020driver}, but these methods can not leverage the natural geographical information (e.g, the connection between lanes) sufficiently.
    
      In general, traffic networks are naturally represented in graph format, where the roads are natural edges and connections between roads act as nodes. In order to overcome the significant limitation of the previous mentioned deep learning methods in traffic flow detection, graph neural networks (GNNs) are applied as an ideal approach to detection problems on traffic networks since spatial dependencies between different nodes have been represented in graph structure. With the input of graph features, variants of GNN architectures have been proposed as the state-of-the-art approach and obtained promising performances in various scenarios \cite{wu2020comprehensive} for detection problem. For instance, Diffusion Convolutional Recurrent Neural Network (DCRNN) \cite{li2018dcrnn_traffic}, Graph Wavenet \cite{wu2019graph} and spatial-temporal Graph convolutional network (STGCN) \cite{DBLP:conf/ijcai/YuYZ18} have been designed to leverage the spatial-temporal information and improve the traditional GNN architecture, which can boost the performance of detection algorithm in highway traffic networks. Tanwi et al. \cite{9413270} refined the DCRNN to transfer the common spatial-temporal information between cities with similar geographical structure to improve the detection performance. Yu et al. \cite{DBLP:conf/ijcai/YuYZ18} proposed STGCN to leverage the spatial and temporal dependencies between different areas of city, to improve the performance of traffic demand forecasting.
    
     Inspired by the successful application of graph modelling, we detect the intention of lane changing based on GNN, in which the graph modelling can extract the spatial information between lanes and enhance the detection performance. Existing works related to detecting lane-change behaviours mainly focus on vehicle-level detection \cite{mandalia2005using, 7835731}. These works try to answer if a specific vehicle has an intention to change lanes while driving on the road to avoid potential collisions. Our work is different in that we aim to detect drivers' lane changing intentions using aggregated information from road-level rather than using the information from individual vehicle. Our focus is to indicate the chaotic level of the current road network so that different levels of traffic intervention may need to be enforced later. Thus, the main objective in our work is not to devise certain strategies to help change drivers' behaviours but to estimate the likelihood of drivers' lane-changing intention using observable information which is not directly collected from individual vehicles but from smart road-side units, e.g., from a camera monitoring system.  
     

	\section{Methodology} \label{methodology}
	
	\subsection{Simulation \& experiment design} 
	
	In this section, the traffic flow data generated by different driving intentions is simulated using SUMO \cite{behrisch2011sumo}. SUMO is open-source software for the simulation of urban mobility, which is prevalent for simulation based studies for intelligent transportation systems. To begin with, we select a specific road segment, i.e., the red segment in Fig. \ref{fig_highway} with length 3.36 km on the motorway M50 in Dublin city, for predicting drivers' lane changing intention. We assume that vehicles will follow VSL on the segment in green but may drive freely as soon as they leave the segment in green, i.e., entering the segment in red shown in Fig. \ref{fig_highway}.
	
\begin{figure}[htp]
    	\centering
    	\includegraphics[width=7.0cm]{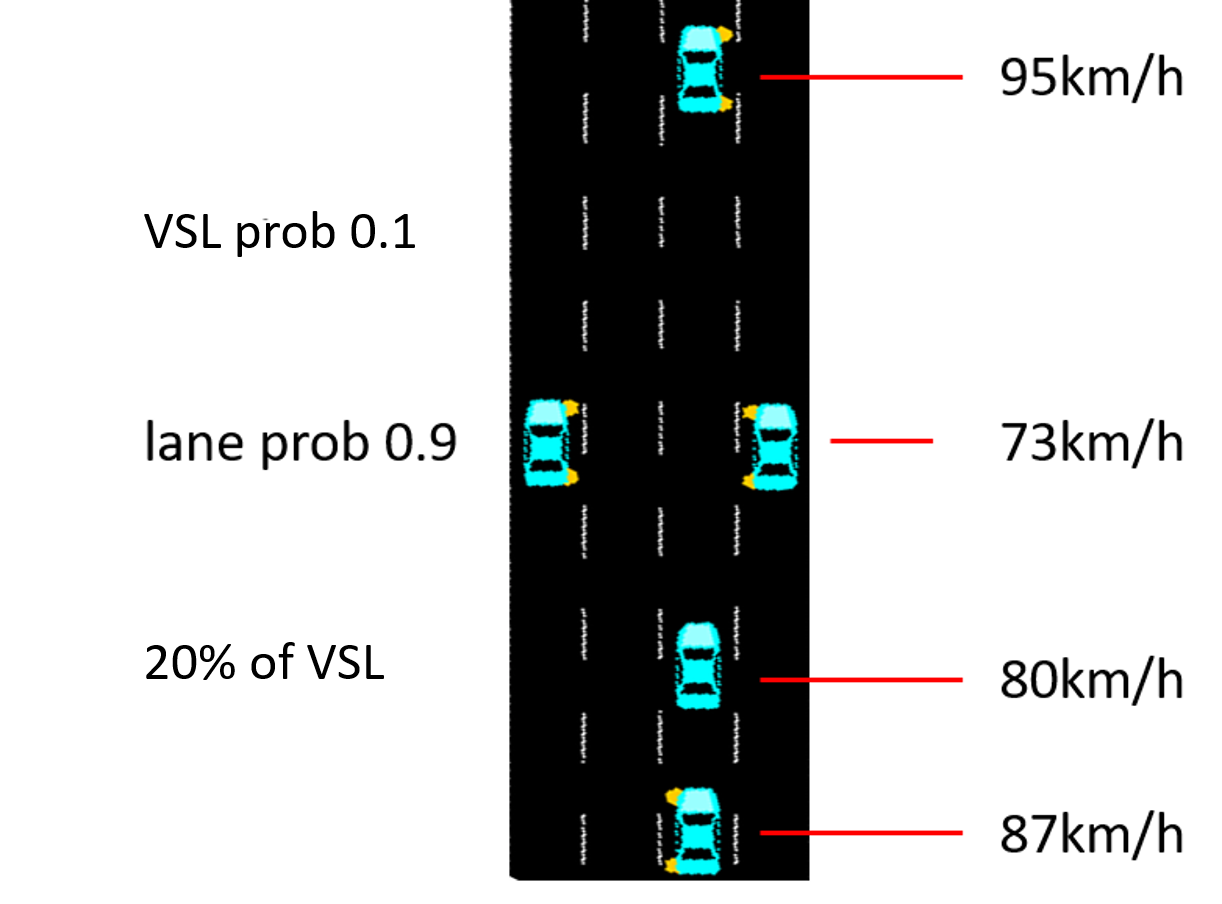}
    	\caption{An example of the setting of driving intention, with setting VSL probability = 0.1, lane probability = 0.9 and $20\%$ of VSL. Each vehicle has 0.1 probability to drive at maximum of VSL (i.e., 80 km/h) or has 0.9 probability to drive between 64 km/h to 96 km/h, and has 0.9 probability to remain in the current lane. The vehicles are driving at a constant speed during the whole journey once their desired speeds are set.}
    	\label{fig_setup}
    \end{figure}

	In our experiment, the car-following model is set in SUMO using a modified version of Krauss model \cite{erdmann2015sumo}. A new vehicle is generated per simulation step (i.e., 1 second) on the lane recommended by SUMO, with safety checks off for driving speed (i.e., speed mode is set as ``32" in SUMO). In a normal situation, all vehicles are driving at maximum speed suggested by VSL without frequent lane changes on the highway traffic network, where the VSL is set as 80 km/h. However, considering that different driving intentions could happen in the real world, we consider the possibility of driving intentions when generating the traffic flow data. Specifically,  driving intentions include violating VSL and frequent lane changing and these intentions may have impacts on traffic flow in our real-world transportation. Violating VSL is defined as driving at a speed that is different from maximum speed of VSL in a given range (e.g, $5\%$, $10\%$, $20\%$ of VSL) and lane changing is defined as the vehicle randomly switches to any lane of the highway traffic network. With this in mind, each vehicle has the possibility (i.e., VSL probability 0.1, 0.5 and 0.9) that drive at maximum of VSL and each vehicle has the possibility (i.e., lane probability 0.1, 0.5 and 0.9) that keep driving at the same lane at each simulation step while driving on the highway traffic network. Fig. \ref{fig_setup} shows a case where VSL probability = 0.1, lane probability = 0.9 and $20\%$ of VSL. This implies that each vehicle has 0.1 probability to drive at maximum of VSL (i.e., 80 km/h) or has 0.9 probability to drive at a speed from 64 km/h to 96 km/h uniformly (i.e., $20\%$ of VSL speed). Each vehicle has 0.9 probability to drive at the current lane (i.e., 0.1 probability changing to other lanes). In a real-world setup, each camera can be set to monitor the movement of vehicles on each lane and access the vehicle number on each lane, e.g., using bounding-box based calculation for each frame. In our setup, we access the average speed and number of vehicles on each lane from SUMO interface to detect drivers' lane change intentions. 
	
	\subsection{Feature selection and model training} 
     
    In our setup, the average driving speed and vehicle number on each lane are collected per simulation step (i.e., 1 second) as features for model training. With this information in place, we intend to estimate the probability of lane changing in scenarios where speed changes of vehicles fall into the following three categories, i.e., $5\%$, $10\%$, $20\%$ of VSL. Therefore, different models for lane changing detection need to be trained with respect to different VSL probability for different range of speed changes. We label the data based on different lane changing intentions (i.e., lane probability). The traffic flow data used for model training, validation and testing are generated for 3600, 1800 and 3600 simulation steps, corresponding to monitoring the traffic flow for 1 hour, a half-hour and 1 hour respectively in real world scenarios.

	\subsection{Traffic flow on graph} \label{graph_modeling}
	
\begin{figure}[htp]
		\centering
		\includegraphics[width=6.0 cm]{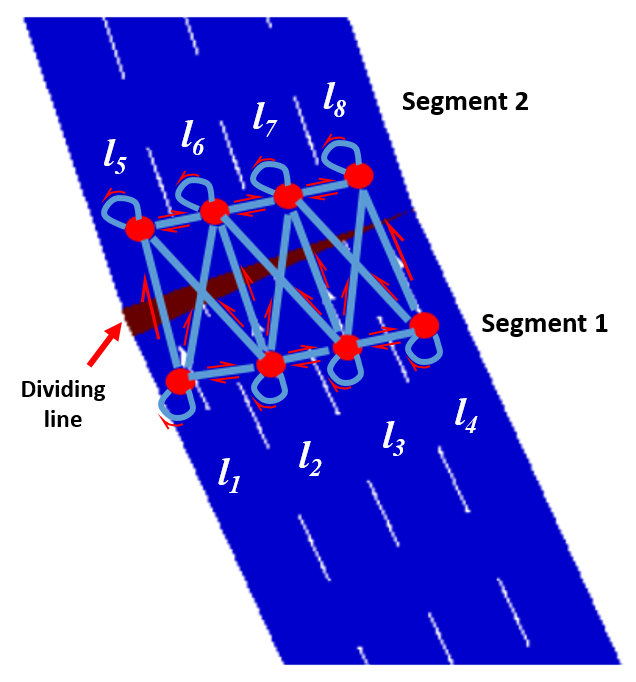}
		\caption{Graph modelling of two consecutive road segment of the M50 highway network. Each lane denotes a node (red points) and the connectivity between different lanes denotes an edge (blue lines). A vehicle can change lane or remain on its current lane depending on its driving intention.}
		\label{fig_graph}
	\end{figure}

	In this section, we model traffic flow data using graph. More specifically, we use $G = (V,E) $ to represent a highway network, where $V$ denotes the nodes representing the set of lane segments $V = \{l_i|i = 1,2,\dots, N\}$, where $N$ denotes the maximum number of nodes in the graph. Let $E$ be edges representing connections between a pair of nodes in the graph $G$. The adjacency matrix is denoted by $A$. The connectivity of the graph is set as fully connected as the vehicle may change lanes from one to any other while driving on the road segment without a VSL system. To better illustrate this point, Fig. \ref{fig_graph} demonstrates our detailed modelling process. The highway network is divided into two road segments with each segment consisting of four lanes, i.e., eight lanes in total $N = 8$. The feature set for lane $i$ at each time step $t$ is denoted by $X_i(t) = (S_i(t),D_i(t))$, where $D_i(t)$ and $S_i(t)$ denote the number of vehicles and the average moving speed on the lane $i$ at time $t$, respectively. Finally, the length of input time window for model training is denoted by $T$ for each of the lane segment.

	\subsection{Network architecture}
	
\begin{figure}[htp]
		\centering
		\includegraphics[width=5.0 cm]{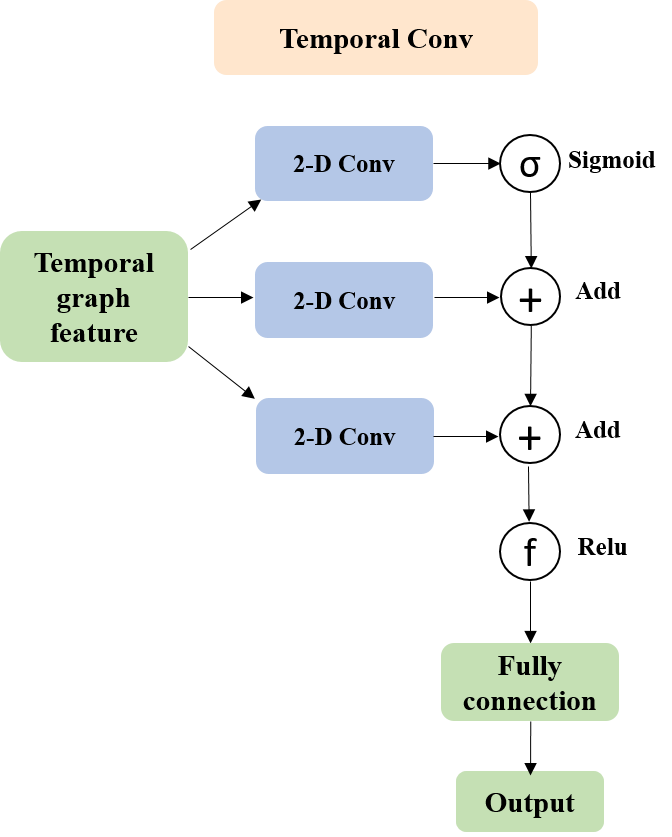}
		\caption{Architecture of temporal convolutional networks. The architecture is refined from \cite{DBLP:conf/ijcai/YuYZ18}.}
		\label{fig_tcnn_network}
	\end{figure}

\begin{figure}[htp]
		\centering
		\includegraphics[width=7.0 cm]{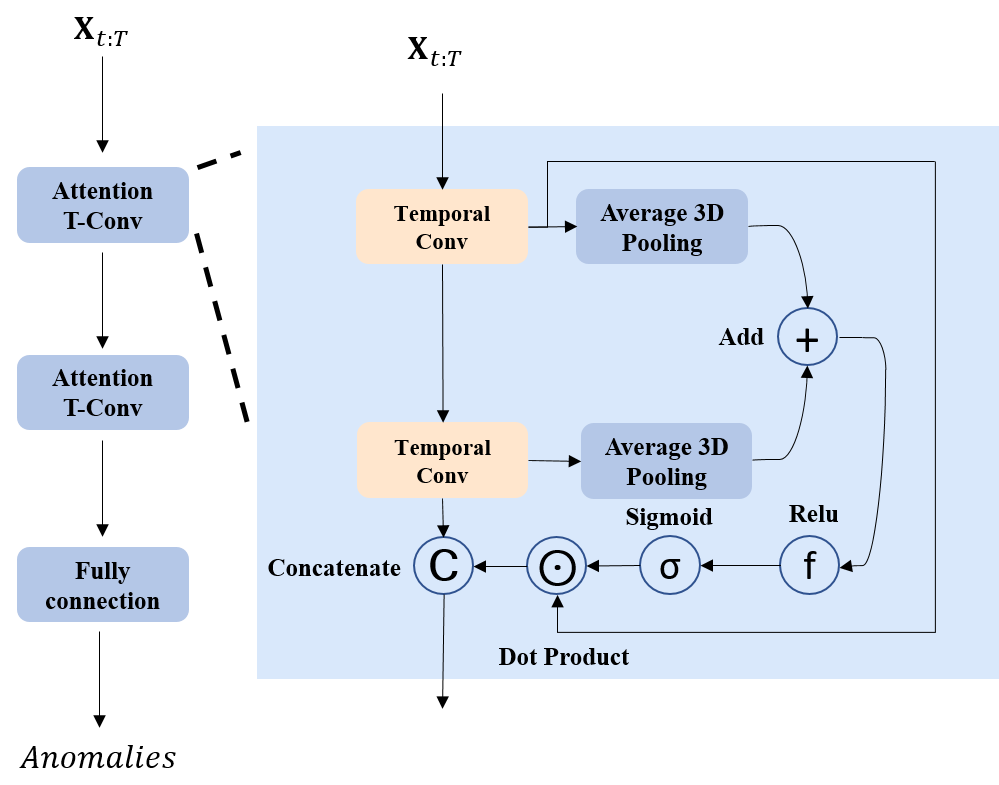}
		\caption{Architecture of attention temporal graph convolutional networks. The architecture is inspired and adapted from  \cite{DBLP:conf/ijcai/YuYZ18, chen2021comparative}. }
		\label{fig_gcn_network}
	\end{figure}

	\begin{itemize}
		
		\item Temporal convolutional neural networks (TCNN). TCNN is designed as a baseline to evaluate the ability of CNN in detecting the intentions given the traffic flow data. The architecture of TCNN is refined from \cite{DBLP:conf/ijcai/YuYZ18} and is demonstrated in Fig. \ref{fig_tcnn_network}. Graph features extracted from each temporal segment are conveyed to three identical 2-D convolutional layers. The output from the first convolutional layer is activated by a sigmoid function to have normalized values between 0 and 1. Output from the other two convolutional layers is added with normalized values and then activated by a Relu function, followed by a fully-connection layer. 
	
		\item Attention-based temporal graph convolutional networks (Lane-GNN). Referring to the work presenting the ST-GCN architecture \cite{DBLP:conf/ijcai/YuYZ18} and built upon our previous work \cite{chen2021comparative}, we introduce TGCN with attention mechanism, consisting of two attention temporal convolution blocks (ATCs) and a fully-connection output layer as our architecture. Each ATC consists of two temporal convolution blocks used in TCNN, with attention mechanism applied to process temporal information, as shown in Fig. \ref{fig_gcn_network}. Note that Lane-GNN has the latent static spatial information since the nodes are fully connected to each other fixedly as aforementioned in Section \ref{graph_modeling}.
	
	\end{itemize}

	\subsection{Network setup}
	
	In our network setup, the number of nodes is 8, indicating 8 lane segments in the considered traffic network. The number of features for each lane is 2, corresponding to the averaged speed and number of vehicles on each lane segment collected as the graph features. The length of temporal segment $T$ is set to 30, 60 and 90 respectively, which will be examined later by our algorithm. For TCNN architecture, each convolution layer has 2 input channels (e.g., corresponding to the number of features) and 64 output channels, with the kernel size as 3.  In each input channel, 2-D traffic data slice with a dimension of [$T$, 8], indicating the specific feature from 8 lanes in a given temporal segment, is used for the model training. Therefore The fully-connection layer receives the input size as [$T$ x 8 x 64] and output size as 3, corresponding to the 3 categories of anomalies that will be discussed in Section \ref{anomalies}.  We set the batch size as 32 indicating there are 32 2-D traffic data slices for each training iteration. The Lane-GNN architecture shares the same setting with TCNN architecture in relation to the temporal convolution layer and fully connection layer. The averaged 3-D pooling operator processes the data along with the dimension of $T$, with the output vector with a size of [1,$T$]. This vector conducts dot product operation with the output of the temporal convolution module, realizing the attention effect on temporal information. Table \ref{tabsettings} summarised common settings when training the Lane-GNN and TCNN.

	\begin{table}[htbp]
		\caption{Network setting for Lane-GNN and TCNN} 
		\begin{center}
			\begin{tabular}{|l|c|c| }
				\hline
				\textbf{Parameters} & \textbf{Values}\\
				\hline
				Nodes  & 8 (only for Lane-GNN)  \\
				Length of temporal segment  & 30, 60, 90\\
				Intention categories & 3 \\
				Feature dimension & 2 \\
				Batch size & 32 \\
				Initial learning rates & 0.001 \\
				Optimizer & Adam algorithm \\
				Weight decay & 0.001 \\
				
				\hline
			\end{tabular}
		\end{center}
		\label{tabsettings}
	\end{table}

	\section{Experimental Results and Discussion} \label{experiment}
	In this section, we present our experimental results and discuss our findings.

	\subsection{Lane changing detection} \label{anomalies}
	
	Here we evaluate the deep learning algorithms for detecting lane changing intentions. In order to exclude the effect of speed violation when detecting the traffic flow caused by lane changing, the data is divided under three conditions, that is data generated under possibilities (i.e., VSL probability  0.1, 0.5 and 0.9) of speed violation. Detection for intentions of lane changing is investigated in these conditions separately. The detection also considers the effect of temporal segments when processing the graph data. We select three temporal segments with different lengths (i.e., $T = 30, 60, 90$) when generating the sample of graph data. With these settings, the algorithm detects the lane changing intention every 30, 60 and 90 seconds respectively in a real-world application. Every two contiguous samples have an overlap time steps of $T/2$ given the specific length of temporal segment $T$.
	
	Table \ref{table_gcn_lane} and Table \ref{table_cnn_lane} demonstrate the results of lane changing detection using Lane-GNN and TCNN respectively. On the one hand, the averaged accuracies based on Lane-GNN are better than that based on TCNN for different ranges of speed change. For each category of speed change, the detection based on Lane-GNN obtains the highest averaged accuracy given the length of temporal segment $T = 90$, which outperforms the performances of TCNN. For instance, Lane-GNN achieves the highest accuracy $99.42\%$ and TCNN obtains accuracy $98.12\%$ given $T = 90$.  On the other hand, the length of a temporal segment is an important factor in our detection task. For Lane-GNN, as the length of temporal segments becomes larger, the accuracy of detecting lane changing behaviours gets better for most conditions regardless of the factor on speed change. For instance, when vehicles have only a 0.1 probability to follow VSL, and the speed changes within the range of $20\%$ of VSL, increasing the length of a temporal segment from $T=30$ to $T=90$ will increase the averaged detection accuracy from $96.06\%$ to $99.13\%$. The average result also aligns well with this finding. In contrast, increasing the length of the temporal segment for TCNN will not have to lead better detection accuracy which can be easily observed from the results reported in Table \ref{table_cnn_lane}. The results from Table \ref{table_gcn_lane} and Table \ref{table_cnn_lane} also indicate that the averaged detection accuracy for both Lane-GNN and TCNN can get better with increasing level of diversity on speed change. However, even in the worst case scenario with $5$\% of VSL, Lane-GNN still shows better averaged detection accuracy ($97.39\%$) than the TCNN method $95.62\%$.

	\begin{table}[h]
		\caption{Detection accuracy using Lane-GNN}
		\label{table_gcn_lane}
		\begin{center}
			\begin{tabular}{|c|c|c|c|c|}
				\hline
				Speed change & Conditions & T=30 & T=60 & T=90\\
				\hline
				5\% of VSL & VSL prob = 0.1 & 90.85\%& 96.00\%& 97.39\%\\
				& VSL prob = 0.5 & 86.62\%& 90.57\%& 94.78\%\\	
				& VSL prob = 0.9 & 98.17\%& 98.86\%& 100.00\%\\
				& Average & 91.88\%& 95.14\%& \textbf{97.39}\%\\
				\hline
				
				10\% of VSL & VSL prob = 0.1 & 91.69\%& 95.71\%& 97.39\%\\
				& VSL prob = 0.5& 90.42\%& 97.14\%& 98.26\%\\
				& VSL prob = 0.9 & 97.61\%& 98.86\%& 97.83\%\\
				& Average & 93.24\%& 97.24\%& \textbf{97.83}\%\\
				\hline
				
				20\% of VSL & VSL prob = 0.1 & 96.06\%& 98.00\%& 99.13\%\\
				& VSL prob = 0.5 & 95.07\%& 98.29\%& 100.00\%\\
				& VSL prob = 0.9 & 96.20\%& 98.86\%& 99.13\%\\
				& Average & 95.78\%& 98.38\%& \textbf{99.42}\%\\

				\hline
			\end{tabular}
		\end{center}
	\end{table}

	\begin{table}[h]
		\caption{Detection accuracy using TCNN}
		\label{table_cnn_lane}
		\begin{center}
			\begin{tabular}{|c|c|c|c|c|}
				\hline
				Speed change & Conditions & T=30 & T=60 & T=90\\
				\hline
				5\% of VSL & VSL prob = 0.1 & 91.27\%& 96.29\%& 90.43\%\\
				& VSL prob = 0.5 & 86.06\%& 91.71\%& 89.57\%\\	
				& VSL prob = 0.9 & 98.59\%& 98.86\%& 97.83\%\\
				& Average & 91.97\%& \textbf{95.62}\%& 92.61\%\\
				\hline
				
				10\% of VSL & VSL prob = 0.1 & 92.54\%& 97.14\%& 94.35\%\\
				& VSL prob = 0.5& 93.52\%& 96.86\%& 97.39\%\\
				& VSL prob = 0.9 & 97.75\%& 98.00\%& 96.96\%\\
				& Average & 94.60\%& \textbf{97.33}\%& 96.23\%\\
				\hline
				
				20\% of VSL & VSL prob = 0.1 & 96.48\%& 98.00\%& 98.26\%\\
				& VSL prob = 0.5 & 95.49\%& 98.00\%& 96.96\%\\
				& VSL prob = 0.9 & 96.76\%& 96.29\%& 99.13\%\\
				& Average & 96.24\%& 97.43\%& \textbf{98.12}\%\\

				\hline
			\end{tabular}
		\end{center}
	\end{table}

	\subsection{Feature visualization and analysis} \label{data_visualization}
	
	In this section, the features (i.e., average speed and vehicle number on the lane) are visualized and the importance of these features are discussed for lane changing detection. 
	
\begin{center}
		\begin{figure*}[ht]
			\centering
			\includegraphics[width= 7.5in, height=2.75in]{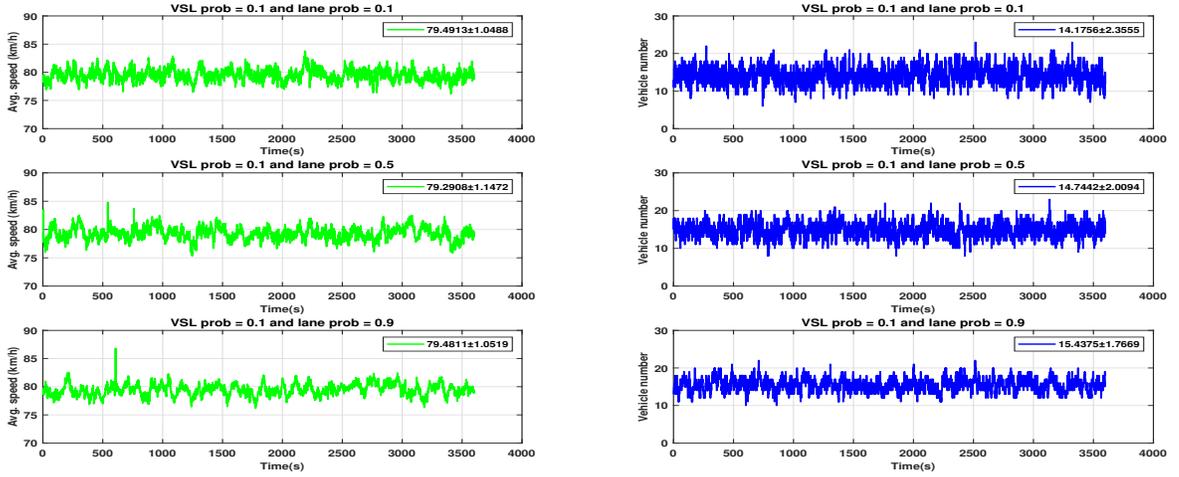}
			\caption{The average speed and vehicle number in lane segment 4 (node $l_4$) among different lane changing probabilities under condition VSL prob = 0.1 and $10\%$ of VSL change. The legends indicate the mean and standard deviation for the corresponding feature.}
			\label{fig_SAS01}
		\end{figure*}	
	\end{center}

\begin{center}
		\begin{figure*}[ht]
			\centering
			\includegraphics[width= 7.5in, height=2.75in]{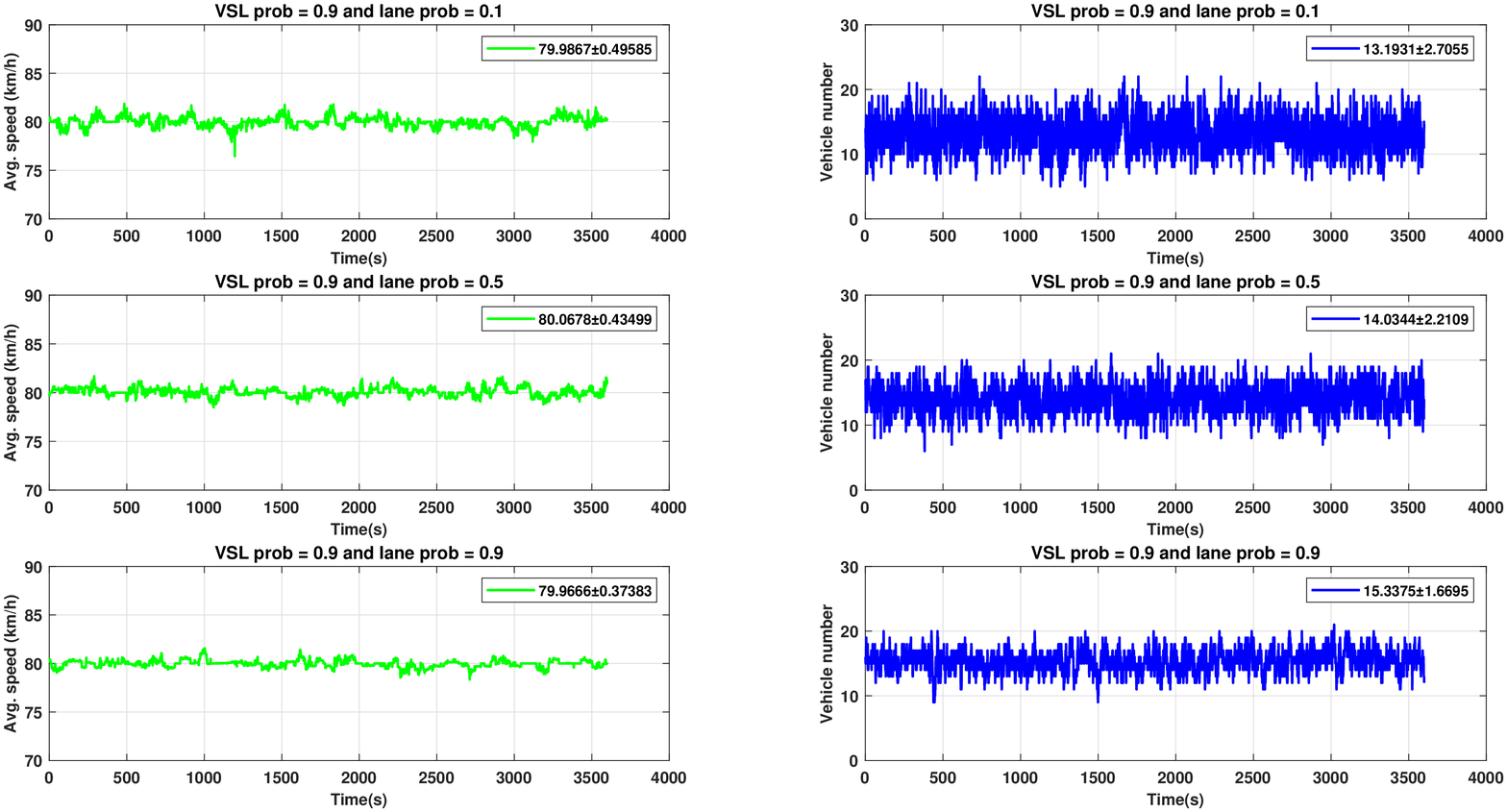}
			\caption{The average speed and vehicle number in lane segment 4 (node $l_4$) among different lane changing probabilities under condition VSL prob = 0.9 and $10\%$ of VSL change. The legends indicate the mean and standard deviation for the corresponding feature.}
			\label{fig_SAS09}
		\end{figure*}	
	\end{center}

	Fig. \ref{fig_SAS01} and Fig. \ref{fig_SAS09} present the averaged speed and vehicle number at lane segment 4 for an hour under different lane changing intentions, where the vehicles have probability of 0.1 and 0.9 to follow the VSL respectively. To better understand these results, Table \ref{Std} shows the statistical mean and standard deviation of the corresponding features. For a given condition on the VSL probability, it is clear to see that the vehicle number feature decreases along with the increasing value of lane probability. However, this observation will not be seen clearly when the average speed feature is under test, e.g., when VSL prob = 0.1. Therefore, the number of vehicle feature contributes more towards the detection accuracy. 
	
	In the spectral domain, the spectral information divergence (SID) measurements \cite{773549} are calculated between averaged speeds and between vehicle numbers. A higher value of SID indicates two signals are more different with respect to the spectrum pattern. Table \ref{SID} shows the SID measurements for averaged speed and vehicle number between different lane probabilities under condition VSL prob = 0.9 and $10\%$ of VSL change. The SID values for the feature of vehicle number are tremendously larger than that for averaged speed, indicating that the vehicle number feature is dominant in our analysis which is consistent with our finding obtained in the time domain analysis.

		\begin{table}[h]
		\caption{Standard deviation for features in different lane probability (LP)}
		\label{Std}
		\begin{center}
			\begin{tabular}{|c|c|c|c|c|}
				\hline
				Conditions & Features & LP0.1 & LP0.5 & LP0.9\\
				\hline
				VSL prob = 0.1 & Avg. Speed & 1.04 & 1.14& 1.05\\
				&Vehicle Num.	& 2.35 & 2.00& 1.76\\	
				\hline
				VSL prob = 0.9 & Avg. Speed & 0.49 & 0.43& 0.37\\
				&Vehicle Num.	& 2.71 & 2.21& 1.67\\	
				\hline

			\end{tabular}
		\end{center}
	\end{table}
	
	\begin{table}[h]
		\caption{Spectral information divergence for features between different lane probability (LP)}
		\label{SID}
		\begin{center}
			\begin{tabular}{|c|c|c|c|}
				\hline
				Features & LP0.1vs.LP0.5 & LP0.1vs.LP0.9 & LP0.5vs.LP0.9\\
				\hline
				Avg. Speed & 46.35 & 49.31& 43.97\\
				Vehicle Num.	& 217.79 & 224.72& 284.41\\	
				\hline

			\end{tabular}
		\end{center}
	\end{table}

	\section{Conclusions} \label{conclusion}

	In this paper, we model the traffic flow data on highway traffic networks using graph and leverage temporal graph convolutional network architecture embedded with attention mechanism to detect the lane changing intentions of drivers. The experiments compared the detection performance of Lane-GNN with that of TCNN. Our results have shown that the attention mechanism enhances the ability in capturing the key temporal information and improves the detection accuracy. With Lane-GNN, lane changing behaviour can be identified within 90 seconds with the highest accuracy of $99.42\%$.  In fact, TCNN is also a promising alternative when a shorter time window, e.g., 30- and 60-second window, is applied, while a longer time window with the Lane-GNN can achieve the best performance overall.
	
	To conclude, we believe that the paper releases implications for future works in lane changing detection for intelligent transportation: 1) graph modelling on traffic flow suits the nature of highway networks and helps to enhance the knowledge representation. 2) The length of the temporal segment affects the performance of anomaly detection. When lane change intentions are required to be detected accurately in important segments of highway traffic networks, delicate models deserve more consideration. On the contrary, if lane changes will not cause severe threats to the driving safety (e.g., in the situation when the driving speed varies within a small range) and can be monitored infrequently, the simpler model can be applied to reduce the computation cost while securing the competitive detection accuracy. 
	
	In future works, advanced feature selection and the effect of components of GNN, such as spatial attention and adaptive adjacency matrix, will be investigated. Finally, we wish to note that the VSL system will cover all parts of M50 highway network in the near future. However, this expansion will not affect the generality of the proposed approach as long as there is a link between a VSL governed road segment and a non-VSL road network, such as M50 to N3/N4 in the Irish road networks. In this context, different attributes of the road segments, e.g., length of lanes, number of lanes, are required to be redesigned for a better modelling of the graph, which forms another part of our future work.
	
	\section*{ACKNOWLEDGMENT}
	This work is supported in part by Science Foundation Ireland (SFI) under Grant Number SFI/12/RC/2289\_P2 (Insight SFI Research Centre for Data Analytics), co-funded by the European Regional Development Fund in collaboration with the SFI Insight Centre for Data Analytics at Dublin City University. The author Hongde Wu is also supported by the research master scholarship funded by the school of electronic engineering, faculty of engineering and computing at Dublin City University. 
	
	\bibliographystyle{ieeetr}
	\bibliography{References}

\begin{thebibliography}{10}

\bibitem{aldegheishem2018smart}
A.~Aldegheishem, H.~Yasmeen, H.~Maryam, M.~A. Shah, A.~Mehmood, N.~Alrajeh, and
  H.~Song, ``Smart road traffic accidents reduction strategy based on
  intelligent transportation systems (tars),'' {\em Sensors}, vol.~18, p.~1983,
  2018.

\bibitem{mejia2021vehicle}
H.~Mejia, E.~Palomo, E.~L{\'o}pez-Rubio, I.~Pineda, and R.~Fonseca, ``Vehicle
  speed estimation using computer vision and evolutionary camera calibration,''
  in {\em NeurIPS 2021 Workshop LatinX in AI}, 2021.

\bibitem{kuvsic2020extended}
K.~Ku{\v{s}}i{\'c}, I.~Dusparic, M.~Gu{\'e}riau, M.~Greguri{\'c}, and
  E.~Ivanjko, ``Extended variable speed limit control using multi-agent
  reinforcement learning,'' in {\em 2020 IEEE 23rd International Conference on
  Intelligent Transportation Systems (ITSC)}, pp.~1--8, IEEE, 2020.

\bibitem{liu2021mpc}
M.~Liu, L.~Cheng, Y.~Gu, Y.~Wang, Q.~Liu, and N.~E. O'Connor, ``Mpc-csas:
  Multi-party computation for real-time privacy-preserving speed advisory
  systems,'' {\em IEEE Transactions on Intelligent Transportation Systems},
  2021.

\bibitem{gu2018design}
Y.~Gu, M.~Liu, M.~Souza, and R.~N. Shorten, ``On the design of an intelligent
  speed advisory system for cyclists,'' in {\em 2018 21st International
  Conference on Intelligent Transportation Systems (ITSC)}, pp.~3892--3897,
  IEEE, 2018.

\bibitem{chen2021intelligent}
B.~Chen, M.~Liu, Y.~Zhang, Z.~Chen, Y.~Gu, and N.~E. O’Connor, ``An
  intelligent multi-speed advisory system using improved whale optimisation
  algorithm,'' in {\em 2021 IEEE 93rd Vehicular Technology Conference
  (VTC2021-Spring)}, pp.~1--6, IEEE, 2021.

\bibitem{liu2015topics}
M.~Liu, {\em Topics in electromobility and related applications}.
\newblock PhD thesis, National University of Ireland Maynooth, 2015.

\bibitem{7350149}
M.~Liu, R.~H. Ordóñez-Hurtado, F.~Wirth, Y.~Gu, E.~Crisostomi, and
  R.~Shorten, ``A distributed and privacy-aware speed advisory system for
  optimizing conventional and electric vehicle networks,'' {\em IEEE
  Transactions on Intelligent Transportation Systems}, vol.~17, no.~5,
  pp.~1308--1318, 2016.

\bibitem{gu2014optimised}
Y.~Gu, M.~Liu, E.~Crisostomi, and R.~Shorten, ``Optimised consensus for highway
  speed limits via intelligent speed advisory systems,'' in {\em 2014
  International Conference on Connected Vehicles and Expo (ICCVE)},
  pp.~1052--1053, IEEE, 2014.

\bibitem{liu2015intelligent}
M.~Liu, R.~H. Ord{\'o}nez-Hurtado, F.~R. Wirth, Y.~Gu, E.~Crisostomi, and
  R.~Shorten, ``An intelligent speed advisory system for electric vehicles,''
  in {\em 2015 International Conference on Connected Vehicles and Expo
  (ICCVE)}, pp.~84--88, IEEE, 2015.

\bibitem{tal2015vehicular}
I.~Tal, B.~Ciubotaru, and G.-M. Muntean, ``Vehicular-communications-based speed
  advisory system for electric bicycles,'' {\em IEEE Transactions on Vehicular
  Technology}, vol.~65, no.~6, pp.~4129--4143, 2015.

\bibitem{xiang2015closed}
X.~Xiang, K.~Zhou, W.-B. Zhang, W.~Qin, and Q.~Mao, ``A closed-loop speed
  advisory model with driver's behavior adaptability for eco-driving,'' {\em
  IEEE Transactions on Intelligent Transportation Systems}, vol.~16, no.~6,
  pp.~3313--3324, 2015.

\bibitem{jeon2014effects}
M.~Jeon, B.~N. Walker, and J.-B. Yim, ``Effects of specific emotions on
  subjective judgment, driving performance, and perceived workload,'' {\em
  Transportation research part F: traffic psychology and behaviour}, vol.~24,
  pp.~197--209, 2014.

\bibitem{behrisch2011sumo}
M.~Behrisch, L.~Bieker, J.~Erdmann, and D.~Krajzewicz, ``Sumo--simulation of
  urban mobility: an overview,'' in {\em Proceedings of SIMUL 2011, The Third
  International Conference on Advances in System Simulation}, ThinkMind, 2011.

\bibitem{huang2014deep}
W.~Huang, G.~Song, H.~Hong, and K.~Xie, ``Deep architecture for traffic flow
  prediction: deep belief networks with multitask learning,'' {\em IEEE
  Transactions on Intelligent Transportation Systems}, vol.~15, no.~5,
  pp.~2191--2201, 2014.

\bibitem{lv2014traffic}
Y.~Lv, Y.~Duan, W.~Kang, Z.~Li, and F.-Y. Wang, ``Traffic flow prediction with
  big data: a deep learning approach,'' {\em IEEE Transactions on Intelligent
  Transportation Systems}, vol.~16, no.~2, pp.~865--873, 2014.

\bibitem{tian2018lstm}
Y.~Tian, K.~Zhang, J.~Li, X.~Lin, and B.~Yang, ``Lstm-based traffic flow
  prediction with missing data,'' {\em Neurocomputing}, vol.~318, pp.~297--305,
  2018.

\bibitem{8526506}
W.~Jiang and L.~Zhang, ``Geospatial data to images: A deep-learning framework
  for traffic forecasting,'' {\em Tsinghua Science and Technology}, vol.~24,
  no.~1, pp.~52--64, 2019.

\bibitem{ma2017learning}
X.~Ma, Z.~Dai, Z.~He, J.~Ma, Y.~Wang, and Y.~Wang, ``Learning traffic as
  images: a deep convolutional neural network for large-scale transportation
  network speed prediction,'' {\em Sensors}, vol.~17, no.~4, p.~818, 2017.

\bibitem{li2016lane}
K.~Li, X.~Wang, Y.~Xu, and J.~Wang, ``Lane changing intention recognition based
  on speech recognition models,'' {\em Transportation Research Part C: Emerging
  Technologies}, vol.~69, 2016.

\bibitem{tang2020driver}
L.~Tang, H.~Wang, W.~Zhang, Z.~Mei, and L.~Li, ``Driver lane change intention
  recognition of intelligent vehicle based on long short-term memory network,''
  {\em IEEE Access}, vol.~8, pp.~136898--136905, 2020.

\bibitem{wu2020comprehensive}
Z.~Wu, S.~Pan, F.~Chen, G.~Long, C.~Zhang, and S.~Y. Philip, ``A comprehensive
  survey on graph neural networks,'' {\em IEEE transactions on neural networks
  and learning systems}, vol.~32, no.~1, pp.~4--24, 2020.

\bibitem{li2018dcrnn_traffic}
Y.~Li, R.~Yu, C.~Shahabi, and Y.~Liu, ``Diffusion convolutional recurrent
  neural network: Data-driven traffic forecasting,'' in {\em International
  Conference on Learning Representations (ICLR '18)}, 2018.

\bibitem{wu2019graph}
Z.~Wu, S.~Pan, G.~Long, J.~Jiang, and C.~Zhang, ``Graph wavenet for deep
  spatial-temporal graph modeling,'' in {\em Proceedings of the 28th
  International Joint Conference on Artificial Intelligence}, pp.~1907--1913,
  2019.

\bibitem{DBLP:conf/ijcai/YuYZ18}
B.~Yu, H.~Yin, and Z.~Zhu, ``Spatio-temporal graph convolutional networks: {A}
  deep learning framework for traffic forecasting,'' in {\em Proceedings of the
  Twenty-Seventh International Joint Conference on Artificial Intelligence,
  {IJCAI} 2018, July 13-19, 2018, Stockholm, Sweden}, pp.~3634--3640,
  ijcai.org, 2018.

\bibitem{9413270}
T.~Mallick, P.~Balaprakash, E.~Rask, and J.~Macfarlane, ``Transfer learning
  with graph neural networks for short-term highway traffic forecasting,'' in
  {\em 2020 25th International Conference on Pattern Recognition (ICPR)},
  pp.~10367--10374, 2021.

\bibitem{mandalia2005using}
H.~M. Mandalia and M.~D.~D. Salvucci, ``Using support vector machines for
  lane-change detection,'' in {\em Proceedings of the human factors and
  ergonomics society annual meeting}, vol.~49, pp.~1965--1969, SAGE
  Publications Sage CA: Los Angeles, CA, 2005.

\bibitem{7835731}
H.~Woo, Y.~Ji, H.~Kono, Y.~Tamura, Y.~Kuroda, T.~Sugano, Y.~Yamamoto,
  A.~Yamashita, and H.~Asama, ``Lane-change detection based on
  vehicle-trajectory prediction,'' {\em IEEE Robotics and Automation Letters},
  vol.~2, no.~2, pp.~1109--1116, 2017.

\bibitem{erdmann2015sumo}
J.~Erdmann, ``Sumo’s lane-changing model,'' in {\em Modeling Mobility with
  Open Data}, pp.~105--123, Springer, 2015.

\bibitem{chen2021comparative}
Z.~Chen, H.~Wu, N.~E. O'Connor, and M.~Liu, ``A comparative study of using
  spatial-temporal graph convolutional networks for predicting availability in
  bike sharing schemes,'' in {\em 2021 IEEE International Intelligent
  Transportation Systems Conference (ITSC)}, pp.~1299--1305, IEEE, 2021.

\bibitem{773549}
C.-I. Chang, ``Spectral information divergence for hyperspectral image
  analysis,'' in {\em IEEE 1999 International Geoscience and Remote Sensing
  Symposium. IGARSS'99 (Cat. No.99CH36293)}, vol.~1, pp.~509--511 vol.1, 1999.

\end{thebibliography}
\end{document}